\pdfoutput=1

\documentclass[letterpaper, 10 pt, journal, twoside]{IEEEtran}

\usepackage{amsmath,amsfonts,bm}

\def\eqref#1{equation~\ref{#1}}

\def\1{\bm{1}}

\def\vmu{{\bm{\mu}}}

\def\vm{{\bm{m}}}

\def\vp{{\bm{p}}}

\def\vx{{\bm{x}}}

\def\mI{{\bm{I}}}

\def\mSigma{{\bm{\Sigma}}}

\DeclareMathAlphabet{\mathsfit}{\encodingdefault}{\sfdefault}{m}{sl}
\SetMathAlphabet{\mathsfit}{bold}{\encodingdefault}{\sfdefault}{bx}{n}

\newcommand{\R}{\mathbb{R}}

\newcommand\cmaes{\mbox{CMA}-\mbox{ES}}

\newcommand\cmamae{\mbox{CMA}-\mbox{MAE}}
\newcommand\cmamegaes{\mbox{CMA}-\mbox{MEGA} (ES)}
\newcommand\cmamegatdes{\mbox{CMA}-\mbox{MEGA} (TD3, ES)}
\newcommand\cmamega{\mbox{CMA}-\mbox{MEGA}}
\newcommand\cmame{\mbox{CMA}-\mbox{ME}}

\newcommand\lmmaes{\mbox{LM}-\mbox{MA}-\mbox{ES}}
\newcommand\lmmamae{\mbox{LM}-\mbox{MA}-\mbox{MAE}}
\newcommand\mapelites{\mbox{MAP}-\mbox{Elites}}

\newcommand\mees{\mbox{ME}-\mbox{ES}}

\newcommand\openaies{\mbox{OpenAI}-\mbox{ES}}
\newcommand\openaimae{\mbox{OpenAI}-\mbox{MAE}}

\newcommand\pgame{\mbox{PGA}-\mbox{ME}}
\newcommand\qdrl{\mbox{QD}-\mbox{RL}}

\newcommand\sepcmaes{\mbox{sep}-\mbox{CMA}-\mbox{ES}}
\newcommand\sepcmamae{\mbox{sep}-\mbox{CMA}-\mbox{MAE}}

\usepackage[group-separator={,}]{siunitx}
\usepackage{xcolor}

\newcommand{\xxnote}[3]{}
\ifx\hidenotes\undefined%
  \renewcommand{\xxnote}[3]{\color{#2}{(#1: #3)}}
\fi

\definecolor{darkgreen}{rgb}{0.13,0.55,0.13}

\newcommand{\hidechange}{} %
\ifx\hidechange\undefined%
  \newcommand{\change}[1]{{\color{blue}{#1}}}
\else
  \newcommand{\change}[1]{#1}
\fi

\newcommand{\hidefinalchange}{} %
\ifx\hidefinalchange\undefined%
  \newcommand{\finalchange}[1]{{\color{blue}{#1}}}
\else
  \newcommand{\finalchange}[1]{#1}
\fi

\newcommand{\hlc}[2][yellow]{{\sethlcolor{#1} \hl{#2}}}

\definecolor{lightgrey}{rgb}{0.9, 0.9, 0.9}

\newcommand{\sref}[1]{Sec.~\ref{#1}}

\newcommand{\fref}[1]{Fig.~\ref{#1}}
\newcommand{\tref}[1]{Table~\ref{#1}}
\newcommand{\aref}[1]{Algorithm~\ref{#1}}

\def\vphi{{\bm{\phi}}}

\usepackage{booktabs}
\usepackage{multirow}
\usepackage{array}
\usepackage{amssymb}
\usepackage{adjustbox}

\newcolumntype{L}[1]
  {>{\raggedright\let\newline\\\arraybackslash\hspace{0pt}}m{#1}}
\newcolumntype{C}[1]
  {>{\centering\let\newline\\\arraybackslash\hspace{0pt}}m{#1}}
\newcolumntype{R}[1]
  {>{\raggedleft\let\newline\\\arraybackslash\hspace{0pt}}m{#1}}

\newcolumntype{X}[2]{%
    >{\adjustbox{angle=#1,lap=\width-(#2)}\bgroup}%
    c%
    <{\egroup}%
}
\newcommand*\rot{\multicolumn{1}{X{90}{1em}}}

\PassOptionsToPackage{hyphens}{url} %
\usepackage{hyperref}
\usepackage{soul} %
\usepackage{caption}
\usepackage{float}
\usepackage{lscape}
\usepackage{graphicx}

\usepackage[ruled,commentsnumbered,linesnumbered,noend]{algorithm2e} %
\DontPrintSemicolon

\title{Training Diverse High-Dimensional Controllers by Scaling Covariance Matrix Adaptation MAP-Annealing
}

\author{Bryon Tjanaka$^{1}$, Matthew C. Fontaine$^{1}$, David H. Lee$^{1}$, Aniruddha Kalkar$^{1}$, Stefanos Nikolaidis$^{1}$%
\thanks{Manuscript received 12 May 2023; accepted 26 August 2023. Date of publication 7 September 2023. This letter was recommended for publication by Associate Editor S. Funabashi and Editor T. Ogata upon evaluation of the reviewers’ comments. This work was supported in part by the NSF CAREER under Grant 2145077, in part by NSF NRI under Grant 2024949, in part by NSF GRFP under Grant DGE-1842487, and in part by NVIDIA Academic Hardware Grant. \textit{(Corresponding author: Bryon Tjanaka.)}}
\thanks{The authors are with the Viterbi School of Engineering, Computer Science Department, University of Southern California, Los Angeles, CA 90007 USA (e-mail: tjanaka@usc.edu; mfontain@usc.edu; dhlee@usc.edu; kalkar@usc. edu; nikolaid@usc.edu).}
\thanks{Source code and videos available at \url{https://scalingcmamae.github.io}}
\thanks{Digital Object Identifier 10.1109/LRA.2023.3313012}
}

\begin{document}

\bstctlcite{MyBSTcontrol}

\maketitle

\begin{abstract}

Pre-training a diverse set of neural network controllers in simulation has enabled robots to adapt online to damage in robot locomotion tasks. However, finding diverse, high-performing controllers requires expensive network training and extensive tuning of a large number of hyperparameters. On the other hand, Covariance Matrix Adaptation MAP-Annealing (CMA-MAE), an evolution strategies (ES)-based quality diversity algorithm, does not have these limitations and has achieved state-of-the-art performance on standard QD benchmarks. However, CMA-MAE cannot scale to modern neural network controllers due to its quadratic complexity. We leverage efficient approximation methods in ES to propose three new CMA-MAE variants that scale to high dimensions. Our experiments show that the variants outperform ES-based baselines in benchmark robotic locomotion tasks, while being comparable with or exceeding state-of-the-art deep reinforcement learning-based quality diversity algorithms.

\end{abstract}


\section{INTRODUCTION}

\IEEEPARstart{B}{y} generating a diverse collection of controllers, we can endow a robot with a variety of useful behaviors.
For example, one popular approach in robotic locomotion has been to train a collection of neural network controllers to enable a walking robot to adapt to damage~\cite{cully2015,colas2020scaling,dqdrl,nilsson2021pga}.
The controllers differ by how often each foot contacts the ground, such that if a foot is damaged, the robot can select a controller that does not rely on that foot.

Searching for diverse controllers may be viewed as a quality diversity (QD) optimization problem~\cite{pugh2016qd}. In QD, the goal is to find solutions $\vphi$ that are diverse with respect to one or more measure functions $m_i(\vphi)$ while maximizing an objective function $f(\vphi)$.
In the locomotion example presented, we search for neural network controller policies $\pi_\vphi$ parameterized by $\vphi$. Each controller should satisfy a unique output of the measure function by using its feet in a different manner from the other controllers, while optimizing the objective by walking forward quickly.

A QD algorithm must balance two aspects given a limited compute budget: \textit{exploring} measure space and \textit{optimizing} the objective.
In our locomotion example, exploration finds new controllers that use the robot's feet a different amount, and optimization makes existing controllers walk faster.

\begin{figure}[t]
\centering
\includegraphics[width=\linewidth]{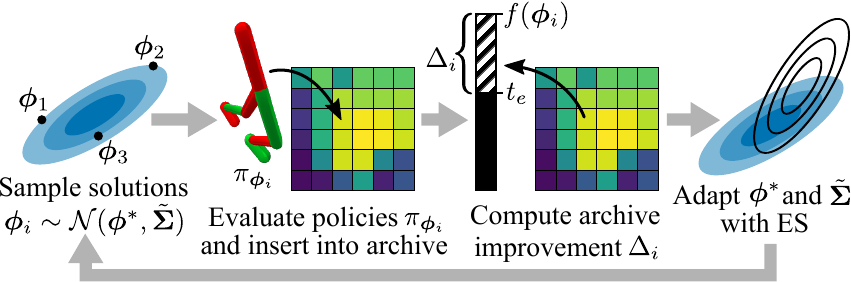}
\caption{We propose variants of the \cmamae{} algorithm that scale to high-dimensional controllers. The variants maintain a Gaussian search distribution with mean $\vphi^*$ and approximate covariance matrix $\tilde{\mSigma}$. Solutions $\vphi_i$ sampled from the Gaussian are evaluated and inserted into an archive, where they generate improvement feedback $\Delta_i$ based on their objective value $f(\vphi_i)$ and a threshold $t_e$ that each archive cell maintains. Finally, the Gaussian is updated with an evolution strategy (ES). Our variants differ from \cmamae{} by incorporating scalable ESs, as the \cmaes{} used in \cmamae{} has $\Theta(n^2)$ time complexity per sampled solution.}
\label{fig:diagram}
\end{figure}

Prior algorithms~\cite{dqdrl,nilsson2021pga} seem to strike a balance between these two aspects of QD, leading to state-of-the-art results. However, these algorithms have practical limitations due to their dependence on deep reinforcement learning (RL) methods. Namely, they must perform time-consuming training of a neural network and have many hyperparameters.

Recent work~\cite{salimans2017evolution} suggests evolution strategies (ES) as a compelling alternative to deep RL methods when optimizing a single controller. Compared with deep RL, ESs do not require network training, and ESs such as the Covariance Matrix Adaptation Evolution Strategy (\cmaes{})~\cite{hansen2016tutorial} are designed to have almost no hyperparameters. Given these benefits, prior work~\cite{colas2020scaling,dqdrl} has developed QD algorithms based on ESs, but these methods have not yet matched the performance of deep RL-based QD methods.

On the other hand, the recently proposed ES-based Covariance Matrix Adaptation MAP-Annealing (\cmamae{}) algorithm~\cite{cmamae} has proven adept at trading off the exploration and objective optimization aspects of QD.
Tuning a single hyperparameter $\alpha \in [0, 1]$ in \cmamae{} enables blending these two aspects, yielding state-of-the-art performance on QD benchmarks.
We hypothesize that \cmamae{}'s ability to balance this tradeoff would enable it to excel in training neural network controllers for robotic locomotion tasks.

\change{While \cmamae{} excels at moderate-dimensional domains, it is intractable for modern neural network controllers because such controllers are \textit{high-dimensional}, i.e., they have thousands or even millions of parameters.
Internally, \cmamae{} guides the QD search with one or more \cmaes{} instances~\cite{hansen2016tutorial}.
Since \cmaes{}'s time complexity is quadratic in the number of parameters, it cannot scale to such controllers.}

\cmaes{}'s complexity arises from how it models the search distribution with a Gaussian that has a full rank $n \times n$ covariance matrix.
However, by replacing this full matrix with sparse approximations, prior work~\cite{largescalecma} %
creates variants of \cmaes{} that scale to high-dimensional problems.

\textit{Our key insight is that we can scale \cmamae{} to high-dimensional controllers by adopting such approximations in its \cmaes{} components}.
Following this insight, we propose three scalable \cmamae{} variants (\sref{sec:scaling}).
To understand their performance and runtime properties, we study these variants on optimization benchmarks (\sref{sec:optbench}).
Next, we evaluate the variants on robotic locomotion tasks (\sref{sec:mainstudy}).
We show that our variants are the highest-performing QD methods based solely on ES.
Furthermore, they are comparable to or exceed the state-of-the-art deep RL-based QD method \pgame{}~\cite{nilsson2021pga} on three of four tasks, while inheriting the aforementioned practical benefits of ES.
We are excited about future applications in other domains,
such as robotic manipulation~\cite{morel2022automatic} and scenario generation~\cite{dsage},
and we have open-sourced our variants in the pyribs library~\cite{pyribs}.

\section{BACKGROUND}

\subsection{Formulation}\label{sec:formulation}

\textbf{Quality diversity (QD).}
Drawing from the definition in prior work~\cite{fontaine2021dqd}, QD considers an objective function $f(\vphi)$ and $k$-dimensional measure function $\vm(\vphi)$,\footnote{It is common to define $\vm(\vphi)$ via $k$ separate measure functions $m_i(\vphi)$. Prior work also refers to measure function outputs as behavior descriptors or behavior characteristics.} where $\vphi \in \R^n$ is an $n$-dimensional solution.
The outputs of $\vm$ form a $k$-dimensional \textit{measure space} $\mathcal{X}$.
The \textit{QD objective} is to find, for every $\vx \in \mathcal{X}$, a solution $\vphi$ such that  $\vm(\vphi) = \vx$ and $f(\vphi)$ is maximized.
Solving this QD objective would require infinite memory since $\mathcal{X}$ is a continuous space, so algorithms based on \mapelites{}~\cite{mouret2015illuminating} relax the QD objective by discretizing $\mathcal{X}$ into a tesselation $\mathcal{Y}$ of $M$ cells.
Then, the QD objective is to maximize the (sum of) objective values of an \textit{archive} $\mathcal{A}$ containing solutions $\vphi_{1..M}$, i.e., $\max_{\vphi_{1..M}} \sum_{i=1}^M f(\vphi_i)$.
Furthermore, $\vphi_{1..M}$ are constrained such that each $\vphi_i$ has measures $\vm(\vphi_i)$ corresponding to a unique cell in $\mathcal{Y}$.

\textbf{Quality diversity reinforcement learning (\qdrl{}).}
As defined in prior work~\cite{dqdrl}, \qdrl{} is a special instance of QD where $\vphi$ parameterizes a reinforcement learning (RL) agent's policy $\pi_\vphi$, and the objective is the agent's expected discounted return in a Markov Decision Process (MDP)~\cite{Sutton2018}.
\qdrl{} also includes a $k$-dimensional measure function $\vm(\vphi)$ that describes the agent's behavior during an episode.

\subsection{Large-Scale Evolution Strategies}\label{sec:es}

An evolution strategy (ES)~\cite{beyer2002} optimizes continuous parameters by adapting a population of solutions such that the population is more likely to attain high performance.
A \textit{large-scale} ES scales to high-dimensional search spaces.

\openaies{}~\cite{salimans2017evolution} is one large-scale ES notable for performing well in RL domains.
It represents a population with an isotropic Gaussian and updates only the Gaussian's mean by passing approximated gradients to Adam~\cite{adam}.

Several large-scale ESs build on Covariance Matrix Adaptation Evolution Strategy (\cmaes{})~\cite{hansen2016tutorial}, an approximate second-order method that
achieves state-of-the-art results in black-box optimization~\cite{hansen2001}.
\cmaes{} models a distribution of search directions with a Gaussian $\mathcal{N}(\vmu, \mSigma)$.
Every iteration, \cmaes{} samples $\lambda$ solutions from this Gaussian and updates it based on rankings of the solutions' performance.

\cmaes{} itself does not scale to high dimensions, as it requires $\Theta(n^2)$ space and $\Theta(n^2)$ runtime per sampled solution.
The space is due to the $n \times n$ covariance matrix $\mSigma$, while runtime stems from two operations.
First, updating $\mSigma$ requires matrix-vector multiplications.
Second, since it is easy to sample from the standard Gaussian $\mathcal{N}(\mathbf{0}, \mI)$ on a computer, sampling from $\mathcal{N}(\vmu, \mSigma)$ is implemented as:
\begin{align}
  \mathcal{N}(\vmu, \mSigma) \sim \vmu + \mathcal{N}(\mathbf{0}, \mSigma) \sim \vmu + \mSigma^{\frac12}\mathcal{N}(\mathbf{0}, \mI) \label{eq:sampling}
\end{align}
The \textit{transformation matrix} $\mSigma^{\frac12}$ requires an $O(n^3)$ eigendecomposition, which \cmaes{} amortizes to $O(n^2)$ per sampled solution by only recomputing $\mSigma^{\frac12}$ every $\frac{n}{\lambda}$ iterations.

Multiple variants~\cite{largescalecma} of \cmaes{} scale to high dimensions by replacing the full covariance matrix with an efficient approximation.
We incorporate \openaies{} and two such variants to scale \cmamae{} to high dimensions.

\subsection{\mapelites{}}

Many QD algorithms, including those in this work, build on \change{Multi-dimensional Archive of Phenotypic Elites} (\mapelites{})~\cite{mouret2015illuminating}.
The vanilla version of \mapelites{} divides the measure space into an archive of evenly-sized grid cells.
Then, it generates solutions by sampling existing solutions from the archive and applying a genetic operator.
These new solutions are inserted into archive cells based on their measures.
If they land in the same cell as a previous solution, they replace the solution only if they have a higher objective.

One recent line of work integrates \cmaes{} into \mapelites{} to optimize for the QD objective (\sref{sec:formulation}).
In Covariance Matrix Adaptation MAP-Elites (\cmame{})~\cite{fontaine2020covariance}, \cmaes{} directly samples solutions, adapting the search distribution to find solutions that create the greatest archive improvement.
\cmame{} runs multiple \cmaes{} instances in parallel, each encapsulated in an \textit{emitter} --- \change{emitters are QD algorithm components that generate solutions for evaluation~\cite{fontaine2020covariance,pyribs}.}
Meanwhile, Covariance Matrix Adaptation \mapelites{} via a Gradient Arborescence (\cmamega{})~\cite{fontaine2021dqd} operates in the differentiable quality diversity (DQD) setting, where exact objective and measure gradients are available.
Here, instead of sampling solution parameters, \cmaes{} branches from a solution point by sampling coefficients that form linear combinations of the objective and measure gradients.

Multiple methods extend \mapelites{} to train neural network controllers, as vanilla \mapelites{} performs poorly in such problems~\cite{colas2020scaling,dqdrl,nilsson2021pga}.
For instance, \cmamega{} cannot be applied to \qdrl{} since it assumes gradients are provided, and such gradients are often unavailable in RL due to non-differentiable environments.
Hence, recent work~\cite{dqdrl} introduces \cmamega{} variants that instead approximate the gradients.
Meanwhile, \mapelites{} with Evolution Strategies (\mees{})~\cite{colas2020scaling} integrates \openaies{} to improve the objective value of a solution point or move the point to a new area of the archive.
Finally, Policy Gradient Assisted \mapelites{} (\pgame{})~\cite{nilsson2021pga} replaces the genetic operator with two operations: (1) gradient ascent, performed with TD3~\cite{fujimoto2018td3}, and (2) crossover, performed with a genetic algorithm~\cite{vassiliades2018line}.
We include these methods as experimental baselines.

\subsection{CMA-MAE}

We extend Covariance Matrix Adaptation MAP-Annealing (\cmamae{})~\cite{cmamae}, a method that builds on \cmame{} and achieves state-of-the-art performance on QD benchmarks.

The key difference between \cmamae{} and \cmame{} is a \textit{soft archive} that enables balancing between optimizing the objective and searching for solutions with new measure values.
This soft archive records a \textit{threshold} $t_e$ for each cell $e$.
$t_e$ is initialized to a minimum objective $min_f$.
When a solution $\vphi$ is inserted into the archive, it is placed into its corresponding cell $e$ if its objective value $f(\vphi)$ exceeds $t_e$.
Then, $t_e$ is updated via polyak averaging $t_e \gets (1 - \alpha)t_e + \alpha f(\vphi)$, where $\alpha \in [0,1]$ is the \textit{archive learning rate}.
Finally, the insertion returns an \textit{improvement value} $\Delta_i \gets f(\vphi) - t_e$, where higher values indicate greater archive improvement.
Note that during insertion, the solution's objective value only needs to cross the threshold, rather than the objective value of the solution previously in the cell.
Thus, implementations must track the best solutions separately, as the archive does not always store them like \mapelites{} does.

Like \cmame{}, \cmamae{} maintains one or more emitters. Each emitter contains a \cmaes{} instance that directly samples solutions from a Gaussian.
By updating the Gaussian based on rankings of the solutions' improvement values, \cmaes{} moves the Gaussian towards solutions more likely to generate high improvement.

The archive learning rate $\alpha$ is a key parameter in \cmamae{}.
When $\alpha = 0$, the threshold remains at $min_f$, so the improvement $\Delta_i$ always equals the objective $f(\vphi)$ (minus a constant $min_f$).
This makes \cmamae{} equivalent to \cmaes{}, as it optimizes solely for the objective.
When $\alpha = 1$, the threshold is equal to the objective value of the solution currently in the cell, which means there is minimal improvement for inserting a solution into a cell with an existing solution.
In this case, \cmamae{} is equivalent to \cmame{}, which always prioritizes discovering new solutions in measure space over improving existing solutions.
Varying $\alpha$ from 0 to 1 smoothly trades off between these two extremes.

\section{SCALING CMA-MAE}\label{sec:scaling}

\begin{algorithm}[t]

\caption{\cmamae{} variants. \hlc[lightgrey]{Highlighted lines} show differences from \cmamae{}~\cite{cmamae}.}
\label{alg:cmamae}

\SetKwProg{CmaMae}{CMA-MAE variants}{:}{}
\CmaMae{$(eval, \vphi_0, N, \psi, \lambda, \sigma, \alpha, min_f)$}{
  \KwIn{$eval$ function that rolls out policy $\pi_\vphi$ and outputs objective $f(\vphi)$ and measures $\vm(\vphi)$, initial solution $\vphi_0$, iterations $N$, number of emitters $\psi$, batch size $\lambda$, initial step size $\sigma$, archive learning rate $\alpha$, minimum objective $min_f$}
  \KwResult{Generates $N\psi\lambda$ solutions, storing elites in an archive $\mathcal{A}$}

  Initialize archive $\mathcal{A}$ and threshold $t_e \gets min_f$ for every cell $e$ \; \label{line:init1}
  Initialize $\psi$ emitters, each with mean $\vphi^* \gets \vphi_0$, \hlc[lightgrey]{covariance matrix approximation $\tilde{\mSigma} \gets \sigma\mI$,} and internal parameters $\vp$ \; \label{line:init2}

  \For{$iter \gets 1..N$}{ \label{line:loop}
    \For{Emitter $1$ .. Emitter $\psi$} { \label{line:emitterloop}
      \For{$i \gets 1..\lambda$} { \label{line:sample}
        \hlc[lightgrey]{$\vphi_i \sim \mathcal{N}(\vphi^*, \tilde{\mSigma})$} \; \label{line:mainsample}
        $f(\vphi_i), \vm(\vphi_i) \gets$ $eval(\vphi_i)$ \; \label{line:eval}
        $e \gets$ calculate\_archive\_cell($\mathcal{A}, \vm$) \;
        $\Delta_i \gets f(\vphi_i) - t_e$ \;
        \If{$f(\vphi_i) > t_e$} { \label{line:checkthresh}
          Replace the solution in cell $e$ of archive $\mathcal{A}$ with $\vphi_i$ \;
          $t_e \gets (1 - \alpha)t_e + \alpha f(\vphi_i)$ \;
        }
      } \label{line:sampleend}
      Rank $\vphi_i$ by $\Delta_i$ \; \label{line:rank}
      \hlc[lightgrey]{Adapt $\vphi^*, \tilde{\mSigma}, \vp$ based on improvement ranking $\Delta_i$} \; \label{line:adapt}
      \If{ES algorithm converges} { \label{line:restart}
        \hlc[lightgrey]{Restart emitter with $\vphi^* \gets$ a randomly selected elite in $\mathcal{A}$, $\tilde{\mSigma} \gets \sigma\mI$, and new internal parameters $\vp$} \;
      } \label{line:restartend}
    } \label{line:emitterloopend}
  } \label{line:loopend}
}

\end{algorithm}

In \cmamae{}, each emitter uses \cmaes{} to update its Gaussian search distribution.
Since \cmaes{} requires $\Theta(n^2)$ space and $\Theta(n^2)$ runtime per solution (with $n$ the solution dimension), \cmamae{} cannot train high-dimensional neural network controllers.
To scale \cmamae{}, we propose three variants that replace \cmaes{} with large-scale ESs.
These variants differ primarily in the complexity of their covariance matrix approximation, and each variant is named by taking its large-scale ES's name and replacing ``ES'' with ``MAE'':
\begin{itemize}
  \item \textbf{\lmmamae{}} substitutes Limited-Memory Matrix Adaptation ES (\lmmaes{})~\cite{lmmaes}, a large-scale \cmaes{} variant that approximates the transformation matrix $\mSigma^\frac12$ with $k \ll n$ $n$-dimensional vectors, each representing a different direction of the search distribution. This rank-$k$ approximation leads to $\Theta(kn)$ complexity.
  \item \textbf{\sepcmamae{}} substitutes Separable \cmaes{} (\sepcmaes{})~\cite{sepcmaes}, a large-scale \cmaes{} variant that constrains the covariance matrix $\mSigma$ to be diagonal, yielding $\Theta(n)$ complexity.
  \item \textbf{\openaimae{}} substitutes \openaies{}~\cite{salimans2017evolution}. As \openaies{} is not a \cmaes{} variant, it differs from \cmaes{} in several mechanisms, but it nevertheless represents the search distribution with a Gaussian, specifically an isotropic Gaussian with constant covariance $\sigma\mI$. Though the covariance is constant, vector operations on the solutions still necessitate $\Theta(n)$ complexity.
\end{itemize}
The listed complexities refer to (1) the space required per emitter, as each emitter maintains its own ES instance, and (2) the runtime required per sampled solution, which is the same regardless of the number of emitters.

\aref{alg:cmamae} and \fref{fig:diagram} show an overview of the variants.
Each variant begins by initializing the archive along with each emitter's ES parameters (line~\ref{line:init1}-\ref{line:init2}). This step includes initializing the covariance matrix approximation $\tilde{\mSigma}$ in lieu of the full covariance matrix $\mSigma$ used in \cmamae{}. Next, each variant repeatedly queries the emitters for solutions (line~\ref{line:loop}). Each emitter (line~\ref{line:emitterloop}) samples $\lambda$ solutions from the distribution $\mathcal{N}(\vphi^*, \tilde{\mSigma})$ (line~\ref{line:sample}-\ref{line:mainsample}). Note that the sampling procedure depends on the approximation employed by the variant.
Once sampled, the solutions are evaluated and inserted into the archive if they cross their cell's threshold $t_e$ (line~\ref{line:eval}-\ref{line:sampleend}). Then, $\vphi^*$, $\tilde{\mSigma}$, and the ES's parameters are updated based on the solutions' improvement ranking (line~\ref{line:rank}-\ref{line:adapt}), such that the emitter is more likely to sample solutions with high improvement on the next iteration. Finally, the emitter restarts if the ES converges (line~\ref{line:restart}-\ref{line:restartend}). We adopt default update and convergence rules from each ES.

\section{OPTIMIZATION BENCHMARKS}\label{sec:optbench}
\begin{table*}[thpb]

\caption{
Optimization benchmark results. We display the metrics described in \sref{sec:optbench_setup} as the mean over 10 trials. QD score is shown as a multiple of $10^6$, and Time is measured in minutes.
}
\label{table:comparison_optbench}

\centering

{
\fontsize{6}{6.9}\selectfont
\setlength{\tabcolsep}{4.7pt}

\begin{tabular}{l rrrr rrrr rrrr rrrr rrrr}
\toprule
{}                 &     \multicolumn{4}{c}{Sphere 100} & \multicolumn{4}{c}{Sphere 1000} &   \multicolumn{4}{c}{Arm 100} &  \multicolumn{4}{c}{Arm 1000}  & \multicolumn{4}{c}{\change{Maze}} \\
\cmidrule(r){2-5}
\cmidrule(r){6-9}
\cmidrule(r){10-13}
\cmidrule(r){14-17}
\cmidrule(r){18-21}
{}                 & QD         & Cov       &          Best &      Time & QD         & Cov       &          Best &      Time & QD         & Cov       &          Best &      Time & QD         & Cov       &         Best & Time      & \change{QD}         & \change{Cov}       &         \change{Best} & \change{Time} \\
\midrule
CMA-MAE            &      0.541 &      0.64 &         98.82 &      2.00 &      0.027 & {\bf0.03} &   {\bf100.00} &    195.60 &      0.787 & {\bf0.79} &    {\bf99.98} &      2.00 &      0.763 &      0.76 &        99.96 &    193.41 &      0.663 &      0.71 & {\bf100.00} &      1.06 \\
LM-MA-MAE          &      0.545 &      0.65 &         99.14 &      1.35 & {\bf0.028} & {\bf0.03} &   {\bf100.00} &     13.42 &      0.784 & {\bf0.79} &    {\bf99.98} &      1.36 & {\bf0.766} & {\bf0.77} &   {\bf99.98} &     13.29 &      0.640 &      0.69 & {\bf100.00} &      0.72 \\
sep-CMA-MAE        & {\bf0.553} & {\bf0.66} &         98.74 &      0.97 & {\bf0.028} & {\bf0.03} &   {\bf100.00} &      3.87 & {\bf0.789} & {\bf0.79} &    {\bf99.98} &      0.98 &      0.763 &      0.76 &        99.96 &      4.18 &      0.741 &      0.79 & {\bf100.00} &      0.47 \\
OpenAI-MAE         &      0.007 &      0.01 &   {\bf100.00} & {\bf0.66} &      0.007 &      0.01 &   {\bf100.00} & {\bf2.67} &      0.770 &      0.78 &         99.97 & {\bf0.79} &      0.714 &      0.72 &        99.96 & {\bf3.19} & {\bf0.751} & {\bf0.81} & {\bf100.00} & \bf{0.38} \\
\bottomrule
\end{tabular}
}

\end{table*}

Replacing \cmaes{} with large-scale ESs in our \cmamae{} variants raises two questions: (1) Since our variants model the search distribution with an approximate Gaussian rather than a full Gaussian, how do they perform relative to each other and relative to \cmamae{}? (2) In practice, are the variants faster than \cmamae{}?
While our goal is to train neural network controllers for robotic locomotion, it is impractical to answer these questions in that domain, since \cmamae{}'s quadratic complexity prevents it from training high-dimensional controllers.
Thus, we first study the variants on lower-dimensional benchmarks.

\subsection{Experimental Setup}\label{sec:optbench_setup}

\textbf{Domains.}
We consider \change{three} QD benchmarks:
(1) In \textit{sphere linear projection}~\cite{fontaine2020covariance}, the objective is the sphere function $f(\vx) = \sum_{i=1}^n x_i^2$, and the measure function linearly projects solutions into a 2D space.
(2) \textit{Arm repertoire}~\cite{vassiliades2018line} considers a planar robotic arm with $n$ equally-sized links. The objective is to find configurations of the $n$ joint angles where the angles have low variance, giving the arm a smooth appearance.
The measures indicate the $x$-$y$ position of the end of the arm.
\change{(3) \textit{Hard maze}~\cite{lehman2011ns} considers a robot that navigates a maze for 250 timesteps.
We use the Kheperax~\cite{kheperax} implementation, where the objective is the robot's energy consumption, and the measures are the final $x$-$y$ position.
The robot is controlled by a neural network with two hidden layers of size 8 and 138 parameters total.
}
In all domains, we linearly transform the objective to the range $[0, 100]$.
We consider 100- and 1000-dimensional versions of sphere and arm, yielding \change{five} domains: \textit{Sphere 100}, \textit{Sphere 1000}, \textit{Arm 100}, \textit{Arm 1000}, \change{\textit{Maze}}.

We select these benchmarks since they are well-studied in the QD literature and exhibit different properties.
For instance, Sphere has a separable objective, and its measure space is intentionally distorted to make it difficult to find new archive solutions.
In contrast, the variance objective in Arm is non-separable, but its measure space tends to be easier to explore, with prior work~\cite{cmamae} showing that even vanilla \mapelites{} fills most of the archive.
\change{Finally, as a small-scale \qdrl{} benchmark, Maze has a less intuitive mapping from neural network parameters to objectives and measures.}

\textbf{Metrics.}
Our primary metric is \textit{QD score}~\cite{pugh2016qd}, which holistically measures algorithm performance by summing the objectives of all archive solutions. To ensure no solution subtracts from the score (this happens if objectives are negative), we subtract the minimum objective (i.e., \cmamae{}'s $min_f$) from all solutions' objectives before computing the score.
Note that $min_f = 0$ in all domains in this section, but $min_f < 0$ in all environments in \sref{sec:mainstudy}.
We also record \textit{archive coverage} (fraction of archive cells containing a solution), \textit{best performance} (highest objective in the archive), and \textit{execution time} (wall-clock time of the experiment).
In our tables, we abbreviate these metrics as ``QD'', ``Cov'', ``Best'', and ``Time.''

\textbf{Procedure.}
In each domain (Sphere 100, Sphere 1000, Arm 100, Arm 1000, \change{Maze}), we conduct a between-groups study with the algorithm (\cmamae{}, \lmmamae{}, \sepcmamae{}, \openaimae{}) as independent variable and the QD score and execution time as dependent variables.
We repeat experiments for 10 trials, where each trial executes an algorithm in a domain for 2 million solution evaluations.
All experiments run on a single CPU core, \change{except in Maze, where we run evaluations on an NVIDIA RTX A6000.}

\textbf{Hyperparameters.}
\cmamae{} and its variants run with archive learning rate $\alpha = 0.001$ (except $\alpha = 0.01$ in Maze) and $\psi = 5$ emitters. Each emitter has batch size $\lambda = 40$ and initial step size $\sigma = 0.02$.
\lmmamae{} sets $k = \lambda = 40$. The Adam optimizer for \openaies{} in \openaimae{} uses learning rate 0.01 and L2 regularization coefficient 0.005.

\textbf{Hypotheses.}
All methods considered model their search distribution with a Gaussian or approximate Gaussian.
We predict that methods with a more complex distribution will perform better but take longer to execute.
To elaborate, the first and simplest algorithm in this ranking is \openaimae{}, which models a fixed isotropic Gaussian.
Since this Gaussian has a constant shape that cannot adapt to the search space, we predict \openaimae{} will have the lowest performance.
However, since \openaimae{} only updates the mean of the Gaussian, it should be the fastest algorithm.

Second, \sepcmamae{} models a diagonal Gaussian.
Since this distribution can change shape and adapt over time, we predict it will lead to higher performance when guiding the QD search.
While the diagonal Gaussian gives \sepcmamae{} the same linear complexity as \openaimae{}, \sepcmamae{} will likely be slower, as it requires additional operations to update the diagonal covariance matrix.

Third, \lmmamae{} uses a rank-$k$ approximation.
While the Gaussian in \sepcmamae{} is limited to being axis-aligned since it is diagonal, the rank-$k$ approximation can represent a more complex Gaussian that is not necessarily axis-aligned.
This property should give \lmmamae{} greater flexibility to adapt to the search space, leading to higher performance.
However, the $\Theta(kn)$ complexity will likely make \lmmamae{} slower than \sepcmamae{}.

Finally, \cmamae{} maintains a full Gaussian, which should be highly flexible and able to adeptly guide the QD search.
The $\Theta(n^2)$ complexity will likely make it the slowest algorithm.
Our hypotheses may be summarized as:

\textbf{H1:} \textit{The QD score will be ranked \openaimae{} \textless{} \sepcmamae{} \textless{} \lmmamae{} \textless{} \cmamae{}.}

\textbf{H2:} \textit{The execution time will be ranked \openaimae{} \textless{} \sepcmamae{} \textless{} \lmmamae{} \textless{} \cmamae{}.}

\subsection{Results}\label{sec:optbench_results}

\tref{table:comparison_optbench} summarizes our results.
To analyze the results, we ran an ANOVA for each dependent variable in each domain.
Before running the ANOVAs, we verified the data were normally distributed through visual inspection and the Shapiro-Wilk test.
Next, we checked homoscedasticity with Levene's test.
In homoscedastic settings, we ran a one-way ANOVA, and in non-homoscedastic settings, we ran Welch's one-way ANOVA.
In almost all domains, we found significant differences across the algorithms for both dependent variables (\tref{table:optbench_anova}).
To analyze the rankings in H1 and H2, we performed pairwise comparisons with Tukey's HSD test or a Games-Howell test, depending on whether the data were homoscedastic or not, respectively.

\begin{table}[thpb]
  \caption{
    One-way and Welch's one-way ANOVA results for each dependent variable on each benchmark.
    All $p$-values are less than 0.005, except for QD Score in Maze.
    Note that large between-group variation led to several high $F$ statistics.
  }
  \label{table:optbench_anova}
  \centering
  {
  \setlength{\tabcolsep}{3pt}
  \fontsize{6}{6.9}\selectfont
  \begin{tabular}{lllll}
  \toprule
  {}             & \multicolumn{1}{c}{QD Score} & \multicolumn{1}{c}{Execution Time} \\
  \midrule
  Sphere 100     & Welch's $F(3, 15.31) = \num{63320}$ & Welch's $F(3, 18.95) = \num{57608}$ \\
  Sphere 1000    & Welch's $F(3, 15.80) = 688.45$      & Welch's $F(3, 15.62) = \num{3.08e6}$ \\
  Arm 100        & $F(3, 36) = 5.46$                   & Welch's $F(3, 18.48) = \num{67102}$ \\
  Arm 1000       & $F(3, 36) = 258.21$                 & Welch's $F(3, 19.10) = \num{2.18e6}$ \\
  \change{Maze}  & \change{$F(3, 36) = \num{0.978}$ ($p > 0.05$)}                       & \change{$F(3, 36) = \num{891.73}$} \\
  \bottomrule
  \end{tabular}
  }
\end{table}

\textbf{H1:}
In all Sphere and Arm domains, \openaimae{} underperformed all other methods. There were no significant differences among the other methods, except that \sepcmamae{} outperformed \cmamae{} in Sphere 100. \change{In Maze, while there was a trend towards \openaimae{} being the best-performing, large variances meant that there were no significant differences among any methods.}

Overall, our results fail to support \textbf{H1}.
Namely, we find that more complex search distributions do not necessarily yield better QD score.
On one hand, as predicted, the most basic distribution (\openaimae{}'s isotropic Gaussian) underperforms \cmamae{} in Sphere and Arm.
\change{However, there is no significant difference between \openaimae{} and \cmamae{} in Maze.
Furthermore, we found no significant differences in any domain between a simple diagonal Gaussian (\sepcmamae{}) and a full Gaussian (\cmamae{}).}

The only case where increasing search distribution complexity increases performance is with \sepcmamae{} outperforming \openaimae{} in Sphere and Arm.
Yet, increasing the complexity further (i.e., \lmmamae{}'s rank-$k$ approximation and \cmamae{}'s full Gaussian) fails to garner further improvement.

\textbf{H2:}
Pairwise comparisons found that the execution time of the algorithms matched the rankings in \textbf{H2}.
The difference between \cmamae{} and the variants was particularly pronounced in the higher-dimensional Sphere 1000 and Arm 1000, where, on average, \cmamae{} took 14.5 times longer than \lmmamae{}, the slowest variant.
These results validate \textbf{H2}, showing that the variants are empirically faster to run than \cmamae{}, and that the variants become faster as their search distribution becomes simpler.

\section{TRAINING HIGH-DIMENSIONAL CONTROLLERS}\label{sec:mainstudy}

We evaluate our \cmamae{} variants' abilities to train diverse, high-performing neural network controllers for robotic locomotion tasks in the QDGym benchmark~\cite{qdgym}.

\begin{table*}[thpb]

\caption{Results from training high-dimensional controllers. We display the corrected metrics described in \sref{sec:mainstudy_setup} as the mean over 10 trials. QD score is shown as a multiple of $10^6$, and Time is measured in hours. \change{We also display the memory usage of each algorithm's components (Space), measured in megabytes. Note that Space is constant across all trials.}}
\label{table:comparison}
\centering

{
\setlength{\tabcolsep}{3.6pt}
\fontsize{6.0}{6.9}\selectfont
\begin{tabular}{lrrrrrrrrrrrrrrrrrrrr}
\toprule
{} & \multicolumn{5}{c}{QD Ant} & \multicolumn{5}{c}{QD Half-Cheetah} & \multicolumn{5}{c}{QD Hopper} & \multicolumn{5}{c}{QD Walker} \\
\cmidrule(r){2-6}
\cmidrule(r){7-11}
\cmidrule(r){12-16}
\cmidrule(r){17-21}
{} &          QD &        Cov &           Best &       Time &   \change{Space} &              QD &        Cov &           Best &       Time &   \change{Space} &          QD &        Cov &           Best &       Time &   \change{Space} &          QD &        Cov &           Best &       Time &   \change{Space} \\
\midrule
LM-MA-MAE          &  {\bf0.761} &       0.36 &       2,297.91 &       7.74 &       112 &           2.830 &       0.62 &       2,243.82 &       6.45 &       87 &       1.135 &       0.54 &       2,845.35 &       4.45 &       77 &       0.327 &       0.42 &       1,289.64 &       5.44 &       84 \\
sep-CMA-MAE        &       0.730 &       0.36 &       2,199.41 &       8.01 &       112 &      {\bf2.892} &  {\bf0.63} &       2,317.81 &  {\bf5.77} &       87 &  {\bf1.173} &       0.55 &  {\bf2,884.33} &  {\bf3.91} &       77 &       0.325 &       0.41 &       1,258.44 &       4.67 &       84 \\
OpenAI-MAE         &       0.638 &       0.34 &       2,073.62 &  {\bf7.25} &       111 &           2.675 &       0.61 &       2,172.15 &       6.30 &       86 &       1.153 &       0.52 &       2,656.07 &       4.51 &       77 &       0.360 &       0.45 &       1,262.87 &       5.67 &       84 \\
PGA-ME             &       0.582 &       0.34 &  {\bf2,711.65} &      18.19 &       374 &           2.682 &       0.56 &  {\bf2,722.17} &      18.19 &      325 &       1.002 &       0.53 &       2,740.55 &      11.89 &      216 &  {\bf0.865} &       0.50 &  {\bf2,685.16} &      12.42 &      291 \\
CMA-MEGA (ES)      &       0.591 &       0.37 &       2,122.80 &       7.43 &  {\bf110} &           2.574 &       0.59 &       2,238.03 &       7.14 &  {\bf85} &       0.502 &       0.53 &       1,508.15 &       3.93 &  {\bf76} &       0.145 &       0.47 &         781.89 &  {\bf3.78} &  {\bf83} \\
CMA-MEGA (TD3, ES) &       0.598 &  {\bf0.40} &       2,349.77 &      14.92 &       374 &           2.693 &       0.59 &       2,495.93 &      18.85 &      325 &       0.935 &       0.52 &       2,498.05 &      11.18 &      216 &       0.789 &  {\bf0.54} &       2,291.46 &      10.96 &      291 \\
ME-ES              &       0.138 &       0.14 &         955.12 &       9.28 &       111 &           0.805 &       0.42 &         848.60 &      10.09 &       86 &       0.185 &       0.42 &       1,043.35 &       4.46 &       77 &       0.037 &       0.30 &         388.58 &       4.32 &       84 \\
MAP-Elites         &       0.444 &       0.38 &       1,160.22 &       7.28 &  {\bf110} &           2.371 &       0.60 &       1,712.17 &       7.09 &  {\bf85} &       0.833 &  {\bf0.56} &       2,420.35 &       4.40 &  {\bf76} &       0.139 &       0.51 &         753.96 &       5.53 &  {\bf83} \\
\bottomrule
\end{tabular}
}

\end{table*}

\subsection{Experimental Setup}\label{sec:mainstudy_setup}

\textbf{Environments.}
\tref{table:envs} shows the QDGym environments considered in this work.
These environments are \textit{unidirectional}, i.e., the objective is to walk forward quickly, and the measures track the proportion of time that each of the robot's feet touches the ground, e.g., if a robot has four legs, it has four measures.
As prior work~\cite{dqdrl} notes, the challenge in these environments arises from \finalchange{performing objective optimization across the entire archive. Namely, it is easy to find a single high-performing controller and fill the rest of the archive with controllers that stand in place and lift their legs to achieve different measures. However, it is difficult to make the robot walk quickly at all points in the measure space.}

As in prior work~\cite{dqdrl,nilsson2021pga}, each domain uses an archive with grid cells.
The robot controller is a neural network mapping states to actions.
The network has two hidden layers of size 128 and tanh activations and is initialized with Xavier initialization.
For the minimum objective $min_f$, QDGym does not have predefined minimum objectives, but we adopt values from prior work~\cite{dqdrl} that recorded the minimum objective inserted into an archive during their experiments.
\tref{table:envs} includes the archive dimensions, number of parameters, and minimum objective in each domain.

\begin{table}[t]
  \fontsize{6}{6.9}\selectfont
  \caption{QDGym locomotion environments~\cite{qdgym}.}
  \label{table:envs}
  \begin{center}
  \begin{tabular}{lrrrr}
    \toprule
    {} &
    \multicolumn{1}{p{0.11\linewidth}}{\centering QD Ant} &
    \multicolumn{1}{p{0.20\linewidth}}{\centering QD Half-Cheetah} &
    \multicolumn{1}{p{0.14\linewidth}}{\centering QD Hopper} &
    \multicolumn{1}{p{0.14\linewidth}}{\centering QD Walker}
    \vspace{4pt}
    \\
    {} &
    \multicolumn{1}{c}{\includegraphics[width=0.12\linewidth]{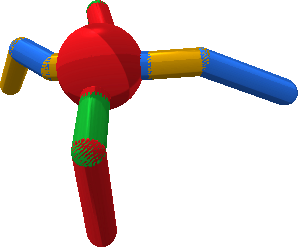}} &
    \multicolumn{1}{c}{\includegraphics[width=0.12\linewidth]{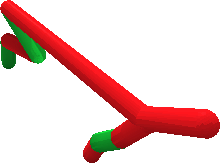}} &
    \multicolumn{1}{c}{\includegraphics[width=0.035\linewidth]{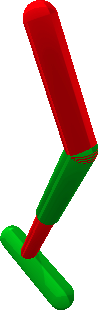}} &
    \multicolumn{1}{c}{\includegraphics[width=0.05\linewidth]{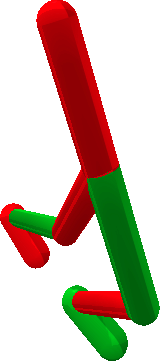}}
    \\
    \midrule
    Archive Grid & [6,6,6,6] & [32,32] & [1024] & [32,32] \\
    Parameters & 21,256 & 20,742 & 18,947 & 20,230 \\
    Min. Objective & -374.70 & -2,797.52 & -362.09 & -67.17 \\
    \bottomrule
  \end{tabular}
  \end{center}
\end{table}

\textbf{Baselines.}
We compare our variants with five baselines: %
\pgame{}~\cite{nilsson2021pga}, two \cmamega{} variants~\cite{dqdrl} that approximate gradients (\cmamegaes{} and \cmamegatdes{}), \mees{}~\cite{colas2020scaling}, and \mapelites{}.
We adopt hyperparameters from the original papers for \pgame{}, the \cmamega{} variants, and \mees{}, except \mees{} uses a population size of 200.
Our \mapelites{} baseline uses isotropic Gaussian noise mutations with standard deviation $\sigma = 0.02$ and batch size 100.
The \cmamae{} variants themselves use the same parameters as in the optimization benchmarks (\sref{sec:optbench_setup}).

\textbf{Procedure.}
We conduct a between-groups study in each environment (QD Ant, QD Half-Cheetah, QD Hopper, QD Walker) with the algorithm (\lmmamae{}, \sepcmamae{}, \openaimae{}, \pgame{}, \cmamegaes{}, \cmamegatdes{}, \mees{}, \mapelites{}) as independent variable and QD score and execution time as dependent variables.
We repeat experiments for 10 trials, where each trial executes an algorithm for 1 million solution evaluations.
Each algorithm runs single-threaded and has 100 CPUs allocated for solution evaluations on a high-performance cluster. %
\change{In addition to these 100 CPUs,} \pgame{} and \cmamegatdes{} are allocated one NVIDIA Tesla P100 GPU to train TD3.

\textbf{Corrected Metrics.}
To save computation, we evaluate each solution for only one episode.
However, unlike the optimization benchmarks, the locomotion environments are stochastic since each episode's initial state is randomly sampled.
Thus, solutions may be inserted into archives due to inaccurate evaluations, e.g., a solution may obtain a high objective by chance. %
Hence, we report \textit{corrected metrics}~\cite{deepgrid,pgamejournal}, where we first re-evaluate all solutions in each final archive for 10 episodes, inserting them into a new, \textit{corrected} archive based on their mean scores.
We then compute the metrics from \sref{sec:optbench_setup} over this corrected archive.

\begin{table*}[thpb]

\caption{Pairwise comparisons for QD score among the variants (for \textbf{H3}) and between the variants and the baselines (for \textbf{H4}) in the locomotion environments.
Each entry compares the method in the row to the method in the column; e.g., \lmmamae{} was significantly better than \openaimae{} in QD Ant.
The symbols used are $<$ (significantly less), $-$ (no significant difference), $>$ (significantly greater), $\varnothing$ (invalid comparison).
We abbreviate \cmamega{} to MEGA for brevity.}
\label{table:pairwise}

\centering

{
\setlength{\tabcolsep}{2.5pt}

\fontsize{6}{6.9}\selectfont
\begin{tabular}{lrrrrrrrrrrrrrrrrrrrrrrrrrrrrrrrr}
\toprule
{} & \multicolumn{8}{c}{QD Ant} & \multicolumn{8}{c}{QD Half-Cheetah} & \multicolumn{8}{c}{QD Hopper} & \multicolumn{8}{c}{QD Walker} \\
\cmidrule(r){2-9}
\cmidrule(r){10-17}
\cmidrule(r){18-25}
\cmidrule(r){26-33}
{} & \rot{LM-MA-MAE} & \rot{sep-CMA-MAE} & \rot{OpenAI-MAE} & \rot{PGA-ME} & \rot{MEGA (ES)} & \rot{MEGA (TD3, ES)} & \rot{ME-ES} & \rot{MAP-Elites} & \rot{   LM-MA-MAE} & \rot{sep-CMA-MAE} & \rot{OpenAI-MAE} & \rot{PGA-ME} & \rot{MEGA (ES)} & \rot{MEGA (TD3, ES)} & \rot{ME-ES} & \rot{MAP-Elites} & \rot{LM-MA-MAE} & \rot{sep-CMA-MAE} & \rot{OpenAI-MAE} & \rot{PGA-ME} & \rot{MEGA (ES)} & \rot{MEGA (TD3, ES)} & \rot{ME-ES} & \rot{MAP-Elites} & \rot{LM-MA-MAE} & \rot{sep-CMA-MAE} & \rot{OpenAI-MAE} & \rot{PGA-ME} & \rot{MEGA (ES)} & \rot{MEGA (TD3, ES)} & \rot{ME-ES} & \rot{MAP-Elites} \\
\midrule
LM-MA-MAE          &       $\varnothing$ &           $-$ &          $>$ &      $>$ &             $>$ &                  $>$ &     $>$ &          $>$ &             $\varnothing$ &           $-$ &          $-$ &      $-$ &             $-$ &                  $-$ &     $>$ &          $>$ &       $\varnothing$ &           $-$ &          $-$ &      $>$ &             $>$ &                  $>$ &     $>$ &          $>$ &       $\varnothing$ &           $-$ &          $-$ &      $<$ &             $>$ &                  $<$ &     $>$ &          $>$ \\
sep-CMA-MAE        &         $-$ &         $\varnothing$ &          $-$ &      $>$ &             $-$ &                  $-$ &     $>$ &          $>$ &               $-$ &         $\varnothing$ &          $-$ &      $-$ &             $-$ &                  $-$ &     $>$ &          $>$ &         $-$ &         $\varnothing$ &          $-$ &      $>$ &             $>$ &                  $>$ &     $>$ &          $>$ &         $-$ &         $\varnothing$ &          $-$ &      $<$ &             $>$ &                  $<$ &     $>$ &          $>$ \\
OpenAI-MAE         &         $<$ &           $-$ &        $\varnothing$ &      $-$ &             $-$ &                  $-$ &     $>$ &          $>$ &               $-$ &           $-$ &        $\varnothing$ &      $-$ &             $-$ &                  $-$ &     $>$ &          $>$ &         $-$ &           $-$ &        $\varnothing$ &      $>$ &             $>$ &                  $>$ &     $>$ &          $>$ &         $-$ &           $-$ &        $\varnothing$ &      $<$ &             $>$ &                  $<$ &     $>$ &          $>$ \\
\bottomrule
\end{tabular}
}

\end{table*}

\textbf{Hypotheses.}
Among the variants, our performance (QD score) prediction remains the same as on the optimization benchmarks (\sref{sec:optbench_setup}),
i.e., the variant with the simplest search distribution (\openaimae{}) will perform worst, and more powerful search distributions (\sepcmamae{} followed by \lmmamae{}) will improve performance.
While the optimization benchmark results (\sref{sec:optbench_results}) did not support this prediction, higher dimensionality in the locomotion environments may highlight differences between the variants.

Compared to the baselines, we believe the smooth improvement ranking in the variants will enable balancing objective optimization and measure space exploration, yielding better performance.
To elaborate, \cmamegaes{} and \cmamegatdes{} use a standard \mapelites{} archive (equivalent to setting $\alpha = 1$ in \cmamae{}'s soft archive), so we think they will focus too much on exploration. %
\pgame{} and \mees{} separate measure space exploration from objective optimization with distinct operations; this separation may be less effective than blending the two aspects.

We predict that execution time differences among the variants will be the same as on the optimization benchmarks; i.e., variants with simpler search distributions will have faster runtimes.
Compared to the baselines, we believe a key factor will be whether a method includes deep RL components.
Unlike the \cmamae{} variants, \pgame{} and \cmamegatdes{} include components of TD3~\cite{fujimoto2018td3} to train actor and critic networks, and these training steps are often time-consuming.
Our hypotheses may be summarized as follows:

\textbf{H3:} \textit{The performances of the \cmamae{} variants will be ordered as \openaimae{} \textless{} \sepcmamae{} \textless{} \lmmamae{}.}

\textbf{H4:} \textit{All \cmamae{} variants will outperform all baselines.}

\textbf{H5:} \textit{The execution times of the \cmamae{} variants will be ordered as \openaimae{} \textless{} \sepcmamae{} \textless{} \lmmamae{}.}

\textbf{H6:} \textit{All \cmamae{} variants will be faster than deep RL-based baselines, i.e., \pgame{} and \cmamegatdes{}.}

\subsection{Results}

\tref{table:comparison} summarizes our results.
Following our analysis procedure in \sref{sec:optbench_results}, we first verified normality through visual inspection and the Shapiro-Wilk test.
Next, Levene's test showed homoscedasticity was violated in all environments.
Thus, we ran Welch's one-way ANOVA (\tref{table:rl_anova}), finding significant differences in all cases.
Finally, we performed pairwise comparisons with the Games-Howell test.

\begin{table}[thpb]
  \caption{
    Welch's one-way ANOVA results in each locomotion environment. All $p$-values are less than 0.001.
  }
  \label{table:rl_anova}
  \centering
  {
  \setlength{\tabcolsep}{3pt}
  \fontsize{6}{6.9}\selectfont
  \begin{tabular}{lllll}
  \toprule
  {}             & \multicolumn{1}{c}{QD Score} & \multicolumn{1}{c}{Execution Time} \\
  \midrule
  QD Ant          & Welch's $F(7, 30.30) = 287.34$  & Welch's $F(7, 29.19) = 54.83$  \\
  QD Half-Cheetah & Welch's $F(7, 30.66) = 207.25$  & Welch's $F(7, 29.55) = 97.66$  \\
  QD Hopper       & Welch's $F(7, 30.33) = 1017.19$ & Welch's $F(7, 30.39) = 60.56$  \\
  QD Walker       & Welch's $F(7, 29.29) = 500.89$  & Welch's $F(7, 29.88) = 111.61$ \\
  \bottomrule
  \end{tabular}
  }
\end{table}

\textbf{H3:} \tref{table:pairwise} shows pairwise comparisons of corrected QD scores for \textbf{H3} and \textbf{H4}.
We find \textbf{H3} unsupported, as there tends to be no significant difference among the variants.
While these results do not align with \sref{sec:optbench_results}'s findings that \openaimae{} often underperforms the other variants, both experiments show that more complex search distributions do not necessarily yield higher performance.

\textbf{H4:} \tref{table:pairwise} shows that the \cmamae{} variants outperform or are not significantly different from prior ES-based methods (\cmamegaes{} and \mees{}), making them the highest-performing ES-based methods in \qdrl{}.
Compared to deep RL-based methods \pgame{} and \cmamegatdes{}, the variants also tend to perform better or have no significant difference.
In particular, both \sepcmamae{} and \lmmamae{} outperform \pgame{} on QD Ant and QD Hopper while having no significant difference in QD Half-Cheetah.
While the variants underperform the deep RL-based methods on QD Walker, prior work~\cite{dqdrl} highlights the importance of deep RL in this task, as only algorithms with TD3 have performed well here.
In short, these results partially support \textbf{H4}, showing that the variants often but not always outperform the baselines.

\begin{table*}[thpb]

\caption{Results from varying the archive learning rate $\alpha$ in \sepcmamae{} in the locomotion environments. We display the corrected metrics described in \sref{sec:mainstudy_setup} as the mean over 10 trials. QD score is shown as a multiple of $10^6$.}
\label{table:comparison_alpha}

\centering

{
\fontsize{6}{6.9}\selectfont

\begin{tabular}{l rrrr rrrr rrrr rrrr}
\toprule
{}                 &     \multicolumn{3}{c}{QD Ant} & \multicolumn{3}{c}{QD Half-Cheetah} &   \multicolumn{3}{c}{QD Hopper} &  \multicolumn{3}{c}{QD Walker} \\
\cmidrule(r){2-4}
\cmidrule(r){5-7}
\cmidrule(r){8-10}
\cmidrule(r){11-13}
{}                 & QD         & Cov       &          Best &  QD         & Cov       &          Best & QD         & Cov       &          Best & QD         & Cov      &         Best \\
\midrule
$\alpha = 0.0$   &       0.314 &       0.13 &  {\bf3,001.76} &           1.577 &       0.48 &       2,106.15 &       0.232 &       0.31 &       1,177.58 &       0.109 &       0.24 &       1,019.89 \\
$\alpha = 0.001$ &  {\bf0.730} &       0.36 &       2,199.41 &           2.892 &       0.63 &       2,317.81 &  {\bf1.173} &       0.55 &  {\bf2,884.33} &       0.325 &       0.41 &       1,258.44 \\
$\alpha = 0.01$  &       0.629 &  {\bf0.41} &       1,914.61 &           2.839 &       0.62 &       2,260.56 &       1.099 &       0.56 &       2,670.70 &  {\bf0.358} &       0.52 &  {\bf1,318.63} \\
$\alpha = 0.1$   &       0.595 &       0.38 &       2,285.24 &      {\bf2.939} &       0.63 &       2,413.25 &       1.054 &  {\bf0.57} &       2,696.78 &       0.318 &  {\bf0.53} &       1,277.05 \\
$\alpha = 1.0$   &       0.387 &       0.31 &       2,132.11 &           2.897 &  {\bf0.63} &  {\bf2,464.52} &       0.647 &       0.55 &       1,964.28 &       0.147 &       0.48 &         853.18 \\
\bottomrule
\end{tabular}
}

\end{table*}

\textbf{H5:} H5 was not supported. We found no significant differences between the variants' runtimes, except \sepcmamae{} was significantly faster than the other variants in QD Walker.
This outcome may arise from the more complex hardware setup of this experiment.
Compared to the single CPU used to run the optimization benchmarks, the evaluations here run on 100 CPUs across multiple nodes.
Slight differences among the nodes may create runtime variance that obscures differences caused by the search distribution complexity.

\textbf{H6:}
All \cmamae{} variants were significantly faster than \pgame{} and \cmamegatdes{} in all domains.
The two deep RL-based algorithms took more than twice as long to run as the variants.
While variance in compute nodes may have contributed to this difference as we believe it did in \textbf{H5}, we believe the majority of the difference stems from the internal algorithm runtime, specifically the aforementioned network training performed in the deep RL-based methods.

\change{
\textbf{Memory Usage:}
To better understand resource requirements, we report the memory usage of each algorithm's internal components in \tref{table:comparison}.
Many algorithms have similar usage due to creating similarly sized components.
For instance, in the \cmamae{} variants, \cmamegaes{}, \mees{}, and \mapelites{}, memory is dominated by the archive, with negligible space for components like emitters.
Meanwhile, \pgame{} and \cmamegatdes{} require more memory to store their TD3 replay buffers.
}

\subsection{Ablation of Archive Learning Rate}\label{sec:ablation}

We believe the soft archive and improvement ranking play a key role in the \cmamae{} variants' performance.
Thus, we ablate this mechanism by varying the archive learning rate $\alpha$ in \sepcmamae{}.
\tref{table:comparison_alpha} shows the result of varying $\alpha \in [0, 1]$; note that all experiments thus far used $\alpha = 0.001$.
These results show that, similar to \cmamae{} in benchmark QD domains~\cite{cmamae}, performance (QD score) falls at the extreme values $\alpha = 0$ and $\alpha = 1$, when \sepcmamae{} focuses entirely on objective optimization or archive exploration, respectively.
In contrast, intermediate values blend both aspects to achieve high performance.

\section{Discussion and Conclusion}

We create variants of \cmamae{} that scale to neural network controllers for robotic locomotion by replacing \cmamae{}'s \cmaes{} component with efficient approximations.
Our results on optimization benchmarks (\sref{sec:optbench}) help distinguish the variants' properties, while our results on locomotion tasks (\sref{sec:mainstudy}) showcase the effectiveness of the variants compared to existing methods.
Furthermore, compared to state-of-the-art deep RL-based methods, our variants bring attractive practical benefits:

\textbf{(1)} The \cmamae{} variants are light on computation. \pgame{} and \cmamegatdes{} both train deep RL components with TD3, a lengthy process that significantly increases runtime as shown in the results of \textbf{H6}.

\textbf{(2)} The \cmamae{} variants have very few hyperparameters since they depend on \cmaes{} and its variants, which are designed to be parameterized by only an initial step size $\sigma$ and batch size $\lambda$. Hence, the \cmamae{} variants only require 5 hyperparameters ($\psi, \lambda, \sigma, \alpha, min_f$, see \aref{alg:cmamae}). \change{In contrast, deep RL-based methods require many more parameters: 18 for \pgame{}, 15 for \cmamegatdes{}. Methods without deep RL require fewer hyperparameters: 5 for \cmamegaes{}, 6 for \mees{}, 2 for \mapelites{}.\footnote{These counts are based on prior listings~\cite{dqdrl} of hyperparameters.} However, our experiments show that such methods do not perform as well as the \cmamae{} variants.}

We emphasize that our \cmamae{} variants are black-box methods that do not leverage the MDP structure of the \qdrl{} problem, making them suitable for settings beyond \qdrl{}.
Hence, we envision future applications of our variants in areas such as manipulation~\cite{morel2022automatic}
and scenario generation~\cite{dsage}.

\addtolength{\textheight}{0cm}

\bibliographystyle{IEEEtran}
\bibliography{references}

\end{document}